\documentclass[conference]{IEEEtran}
\IEEEoverridecommandlockouts
\usepackage{cite}
\usepackage{amsmath,amssymb,amsfonts}
\usepackage{graphicx}
\usepackage{booktabs}
\usepackage{multirow}
\usepackage{algorithm, algorithmic}
\usepackage{placeins}
\usepackage{textcomp}
\usepackage{xcolor}
\usepackage{pgfplots}
\pgfplotsset{compat=1.18}

\definecolor{CallistoBlue}{HTML}{0072B2}
\definecolor{CallistoGreen}{HTML}{009E73}
\definecolor{CallistoRed}{HTML}{D55E00}
\definecolor{CallistoGrid}{HTML}{D9D9D9}
\definecolor{CallistoAxis}{HTML}{555555}

\pgfplotsset{
  callisto axis/.style={
    width=0.78\linewidth,
    height=0.58\linewidth,
    scale only axis,
    clip=false,
    axis background/.style={fill=white},
    axis line style={draw=CallistoAxis, line width=0.35pt},
    tick style={draw=CallistoAxis, line width=0.35pt},
    major grid style={draw=CallistoGrid, line width=0.35pt},
    grid=major,
    font=\footnotesize,
    tick label style={font=\scriptsize},
    label style={font=\footnotesize},
    title style={font=\small\bfseries, yshift=-1pt},
    legend style={
      font=\scriptsize,
      draw=CallistoGrid,
      fill=white,
      fill opacity=0.92,
      text opacity=1,
      cells={anchor=west},
      row sep=-1pt
    },
    every axis plot/.append style={line width=1.05pt},
    /pgf/number format/1000 sep={\,}
  },
  callisto labeler line/.style={
    color=CallistoBlue,
    mark=*,
    mark size=1.55pt,
    mark options={fill=CallistoBlue, draw=white, line width=0.25pt}
  },
  callisto magika line/.style={
    color=CallistoGreen,
    densely dashed,
    mark=diamond*,
    mark size=1.75pt,
    mark options={fill=CallistoGreen, draw=white, line width=0.25pt}
  },
  callisto labeler fill/.style={
    mark=none,
    draw=white,
    fill=CallistoBlue,
    fill opacity=0.55
  },
  callisto magika fill/.style={
    mark=none,
    draw=white,
    fill=CallistoGreen,
    fill opacity=0.55
  }
}

\usepackage{url}
\usepackage{float}
\usepackage{hyperref}
\hypersetup{hidelinks}

\def\BibTeX{{\rm B\kern-.05em{\sc i\kern-.025em b}\kern-.08em
    T\kern-.1667em\lower.7ex\hbox{E}\kern-.125emX}}

\begin{document}

\title{Fast And Accurate Text Content File Type Identification}

\author{
\IEEEauthorblockN{Manu Nandan}
\IEEEauthorblockA{
\textit{CrowdStrike, Inc.}, USA\\
manu.nandan@crowdstrike.com
}
\and
\IEEEauthorblockN{Michael Brautbar}
\IEEEauthorblockA{
\textit{CrowdStrike, Inc.}, USA\\
michael.brautbar@crowdstrike.com
}
\and
\IEEEauthorblockN{Edward Raff}
\IEEEauthorblockA{
\textit{CrowdStrike, Inc.}, USA\\
edward.raff@crowdstrike.com\\
\textit{Univ. of Maryland, Baltimore County}, USA\\
raff.edward@umbc.edu
}
}

\maketitle

\begin{abstract}
A common requirement across organizations is to have a tool that can identify file types based on their contents, particularly in the cybersecurity domain where magic numbers and file extensions can not be trusted. While existing tools work well in practice, there is plenty of room for improvement either in terms of computational load and time for detection in the case of model based tools like Magika \cite{fratantonio25_magika} or in terms of accuracy of detection in the case of file parsing tools that use programming language constructs. In this study, we propose a neural network model for identification of types of text content files, especially source code, that is more accurate and faster than other available tools. Our experiments on open-source files indicate that it is not only more accurate on average for text-content file-type identification, but also approximately four times faster than Magika, while being 28\% smaller in size.
\end{abstract}

\maketitle

\section{Introduction} \label{sec:intro}

A wide variety of tools has been developed over the years to identify file types based on their contents, a ubiquitous need in common computer applications. Multiple domains, ranging from web applications to cybersecurity, have a critical need for such a tool. In the cybersecurity domain, file type detection has several applications such as malware detection, policy enforcement, and data protection. Depending on the identified file type, there might be multiple actions taken such as (1) invoking another model based or heuristic mechanism to determine if it is malicious and blocking the file execution or logging it, or (2) prevent ex-filtration attempts by blocking attempts to upload files to the internet if they are in one of the programming languages that an organization uses for development. Such actions are crucial to maintain a good security posture, and the use of these tools is one of the preliminary steps that are commonly taken in cybersecurity solutions. The same concerns also apply to the usage of such tools in web applications, where only some file types are allowed to be uploaded to prevent the spread of malware. Other applications include software development tools (for e.g., Visual Studio Code) for syntax highlights and web browsers to decide how to render file content in different formats.

Considering the security-focused applications of these tools, it is important to prevent simple methods of evasion of detection, such as by renaming file extensions, by designing them to identify the file types from the file contents. Given the volume of files and their ever-increasing size, the tools need to be very quick to detect file type while having high detection accuracy. If the detection time is more than a few milliseconds, the end user will have a bad experience and risks administrators and users disabling this important functionality. In addition, if such methods are computationally complex, they will again adversely affect end users' experience by imposing high computational load when running the detection tools, thereby consuming all CPU resources.

The existing solutions are typically either model-based or signature-based. An example of a model-based solution is Magika \cite{fratantonio25_magika}, an open-source tool reported to achieve state-of-the-art performance in not only text content file type identification but also binary file type identification. Signature-based file-type identification, though widely used and computationally lightweight, has been found to perform relatively poorly in practice for text-file-type identification. Specifically, the areas of weakness of the current approaches in general are the following:
\begin{itemize}
    \item Low accuracy of file type identification for text content file types, especially for signature-based methods. Applications such as cybersecurity require very high accuracy, without which downstream systems can fail.
    \item High computational requirements for model-based methods. It is not uncommon for a large number of files to require simultaneous file-type identification, for example, when files are uploaded to a website. Fast file type identification methods with low memory and CPU footprints are a necessity in these situations to ensure good user experience.
    \item Evaluation on proprietary data. Evaluation of methods on publicly available data would facilitate comparison and benchmarking.
\end{itemize}
Magika, in particular, was found to yield high accuracy across most of our use cases. However, it was slower than needed, with it taking close to 4 milliseconds to identify a single file's type even while using multiple CPUs. In addition, multiple file types that needed to be identified were not supported, e.g., AppleScript and ColdFusion. In this study, we propose \textsc{Labeler}, a neural network model designed specifically for files with text content, such as source code, configuration file formats and markup languages, that is more accurate and faster than Magika. To efficiently process text content files, we also propose a custom lightweight tokenizer that converts file contents into integer tokens for consumption by the model. As validated in our experiments, \textsc{Labeler} achieves a Macro $F1$ that is 8\% higher than Magika, with individual file-types often 10 to 20 percentage points better, while being 28\% smaller and running close to 4 times faster, a Pareto improvement.

\section{Related Work} \label{sec:related_work}

The other methods for file type identification can be broadly classified into two groups: (1) Signature-based and (2) Model-based, as described below.

\subsection{Signature-Based File Type Identification}
For text content file types, signature-based identification uses regular expressions to detect common usage patterns. The utility \textsc{file} \cite{file_nix,libmagic} is used in all BSD and Linux distributions. Similarly, Apache Tika \cite{ApacheTika} uses parsers to identify a limited set of text content file types. These tools and others exiftool \cite{exif} and trid \cite{trid} identify a wide variety of binary file types using ``Magic Bytes'', Multipurpose Internet Mail Extensions (MIME) types or other file metadata. Magic bytes are predefined byte sequences that exist at a specific location of the file and are generally found to be good indicators. Utilities such as file \cite{file_nix} are reported to be fast and accurate in binary file type identification \cite{fratantonio25_magika}. However, they are easy to adversarially subvert by modifying the magic bytes, making them inappropriate to use in a security context.

Such methods, though computationally efficient, suffer from low detection accuracy and lack of adaptability for text content file types. The regular expressions used for file type identification need periodic updates as usage patterns evolve and new features are added to languages. Hence, there is a need for highly manual work from subject matter experts. Model-Based methods can automatically learn the usage patterns most relevant to distinguish between file types if trained with large and updated training sets, so that not only are they more accurate, but also require less manual effort. With this motivation, \textsc{Labeler} was designed to be a model-based method.

\subsection{Model-Based File Type Identification} \label{sec_related_work_model_based}
Several approaches have been studied over the years for identifying file types from their contents. Guesslang \cite{guesslang} is a neural network model based on a wide and deep network architecture \cite{Cheng_16}. It was trained on 1.9 million open-source code files and processes file content by computing bi-grams from the entire file. It is used in popular applications, including Visual Studio Code \cite{guesslang_VS_code}. It supports the identification of fifty-four commonly used source code file types. The original guesslang implementation is in a deprecated TensorFlow version, but Visual Studio Code developers maintain a version that has a javascript interface \cite{vscode_guesslang}.

\begin{figure}[!h]
    \centering
    \includegraphics[width=0.85\linewidth]{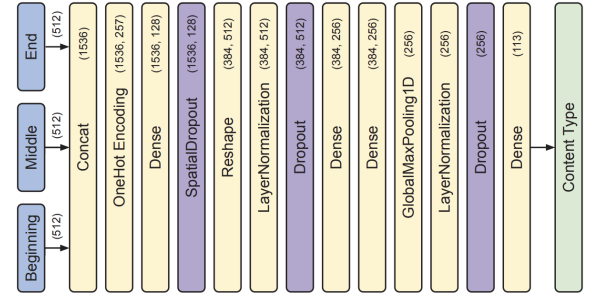}
    \caption{Magika model architecture from \cite{fratantonio25_magika}. It works on three byte sequences extracted from the start, middle, and end of the files. The numbers next to the layers’ names indicate the size of their outputs.}
    \label{fig:magika}
\end{figure}

Magika \cite{fratantonio25_magika}, is reported to have state of the art performance in the identification of more than two hundred file types covering both binary and text types. It is a deep neural network model that works directly on byte sequences sliced from the file contents. Specifically, it works on three chunks of byte sequences extracted from the start, middle, and end of the file. The classifier model works on the three sequences of byte (i.e., three regions extracted from the file content) to identify the file type. The model is relatively small, with parameters reported to be approximately 1 MB.  Magika has been integrated into popular applications such as Google Drive and Gmail. It is also integrated into Apache Tika \cite{tika_magika}. The model architecture is shown in Figure \ref{fig:magika}. 

Some approaches \cite{FITZGERALD2012S44,Wang_18} use NLP-based feature extraction with Support Vector Machines or deep learning models \cite{Mittal_21, Kristian_23} to address the file-fragment classification problem (i.e., you have an incomplete file, often recovered from a corrupted storage or data stream) in digital forensics and file recovery~\cite{McDaniel:2003:CBF:820756.821828,Pal2009,Roussev:2009:FFC:1683311.1684973,Axelsson2010S24,Gopal2011,Poisel2013}. These methods are reported to have lower detection performance \cite{fratantonio25_magika} than Magika and hence are not widely used outside of forensic contexts where they are truly necessary. 
Beyond having to deal with partial file fragments, digital forensics often relies on more complex and heavy-lift processing techniques to try to mitigate the impact on accuracy. This includes creating digests based on compression methods, entropy, and other statistics~\cite{raff_lzjd_digest,chang_fbhash_2019,Axelsson2010S24,raff_lzjd_2017}, complex domain knowledge parsing in the face of incomplete data~\cite{Clemens2015,Hand2012,vandenBos:2011:BDL:1985793.1985887}, and neural networks~\cite{Wilgenbus2013,Raff2020b,MalConv}, among many machine learning approaches~\cite{Sportiello2011,Gopal2011}.
\cite{Xiong_bert_2024} proposed using a fine-tuned BERT model for file-type identification. While such approaches could possibly yield good accuracy in file type detection, they suffer from high computational load for inference and are not suitable for the applications described in Section \ref{sec:intro}.

A thorough evaluation of guesslang is presented in \cite{fratantonio25_magika} along with comparisons against signature-based tools such as \textsc{file}, exiftool and trid. For identification text content file types, guesslang was reported to have an average F1 score of 0.77, which is higher than file, exiftool or trid. However, it was significantly lower than the average F1 score of 0.99 of Magika for the same file types. In this study, we used only Magika as a baseline, given its reported state-of-the-art performance by a very large margin compared to the other tools, and its widespread usage. As described in the following sections, \textsc{Labeler} is designed specifically for text content file types with the use of a custom tokenizer, unlike Magika. It has a custom neural network architecture designed for efficiency and is $3.7\times$ faster despite processing $5.3\times$ more input and using one CPU core instead of multiple.

\section{Method} \label{sec:method}
Given the need for a lightweight, high-accuracy model to specifically identify text content types, the proposed model uses a simple tokenizer described in the next section. Similar to the method in \cite{fratantonio25_magika}, file contents are read in chunks. However, our solution reads only two chunks of a given file, rather than three like in Magika. The first 4 Kb and the last 4 Kb of the file are processed as described in Section \ref{sec: tokenizer}. The tokenizer converts the text content in each file chunk (from start to end) into a sequence of integers, called tokens, based on a mapping of text substrings to specific integer values. 
The two token sequences computed by the tokenizer are processed by the neural network architecture described in Section \ref{sec:model_arch}, to train the model. This enables the model to predict the file type in a fixed time irrespective of file size. 4 Kb is selected from the start/end of the file because the start/end often have file-specific meta data or mandatory structures that aid in detection, while also being large enough to extend past such mandatory structures and obtain a large sample of non-structural content. 

\subsection{Tokenizer} \label{sec: tokenizer}

Each text content file undergoes tokenization, a process in which text is converted into integer sequences for computational processing. Our custom tokenization methodology is formalized in Algorithm~\ref{alg:tokenization} described below. We designed the tokenizer with the goal of capturing important symbols in the files without being computationally complex.

\begin{algorithm} [h!]
\caption{Text Content Symbol Tokenization}
\label{alg:tokenization}
\begin{algorithmic}[1]
\REQUIRE Text content file $F$
\ENSURE Token sequence $T$
\STATE Split $F$ into lines $L = \{l_1, l_2, \ldots, l_n\}$
\FOR{each line $l_i \in L$}
    \STATE Remove leading and trailing whitespace from $l_i$
    \STATE Split $l_i$ into sub-words $S$ separated by non-alphanumeric characters
    \FOR{each sub-word $s_j \in S$}
        \IF{$s_j$ is alphanumeric}
            \STATE Split $s_j$ at case transitions (lowercase to uppercase)
            \STATE Convert all alphanumeric components to lowercase
        \ENDIF
        \STATE Append processed tokens to $T$
    \ENDFOR
\ENDFOR
\RETURN $T$
\end{algorithmic}
\end{algorithm}

For example, the code fragment \texttt{if len(sub\_part) == camelPart} undergoes the following transformation: initial splitting yields \texttt{if}, \texttt{len}, \texttt{(}, \texttt{sub}, \texttt{part}, \texttt{)}, \texttt{==}, \texttt{camelPart}; subsequent case-based segmentation of \texttt{camelPart} produces \texttt{camel}, \texttt{part}; and final normalization results in the token sequence: \texttt{if}, \texttt{len}, \texttt{(}, \texttt{sub}, \texttt{part}, \texttt{)}, \texttt{==}, \texttt{camel}, \texttt{part}.

This tokenization approach is designed to preserve relevant symbols in source code and configuration files, distinguishing it from conventional natural language processing tokenizers, which typically discard punctuation or employ word-piece segmentation. Such traditional methods prove inadequate for our analysis. For instance, while the model must differentiate between \texttt{define} and \texttt{def} as distinct tokens, a word-piece tokenizer \cite{Schuster2012JapaneseAK} may fragment \texttt{define} into \texttt{def}, \texttt{in}, and \texttt{e}, thereby obscuring their semantic distinction. 

Although the tokenizer accepts inputs of up to 4 KB, the resulting token sequences may vary in length. This is due to the fact that sub-words computed from the input text will have varying lengths, and depending on the nature of sub-words in the input, the number of tokens in the output varies. This is handled by truncating the output of the tokenizer to 512 tokens. In the event that the number of computed tokens is less than 512, the token sequence will be zero-padded as is general practice in neural networks.

\subsubsection{Vocabulary Construction}

The processed token strings are converted into integer representations through the following procedure:
\begin{enumerate}
    \item Compute the frequency of each symbol or sub-word across the training corpus.
    \item Rank sub-words by occurrence frequency in descending order.
    \item Select the $K$ most frequent symbols to form the vocabulary.
    \item Map each symbol to its corresponding rank.
\end{enumerate}
Symbols not present in the vocabulary are represented as unknown tokens. Empirical evaluation revealed that a relatively compact vocabulary size of $K = 4096$ is sufficient to capture the lexical diversity of programming languages without compromising model performance.

\subsection{Model Architecture}\label{sec:model_arch}

\begin{figure}
    \centering
    \includegraphics[width=0.55\linewidth]{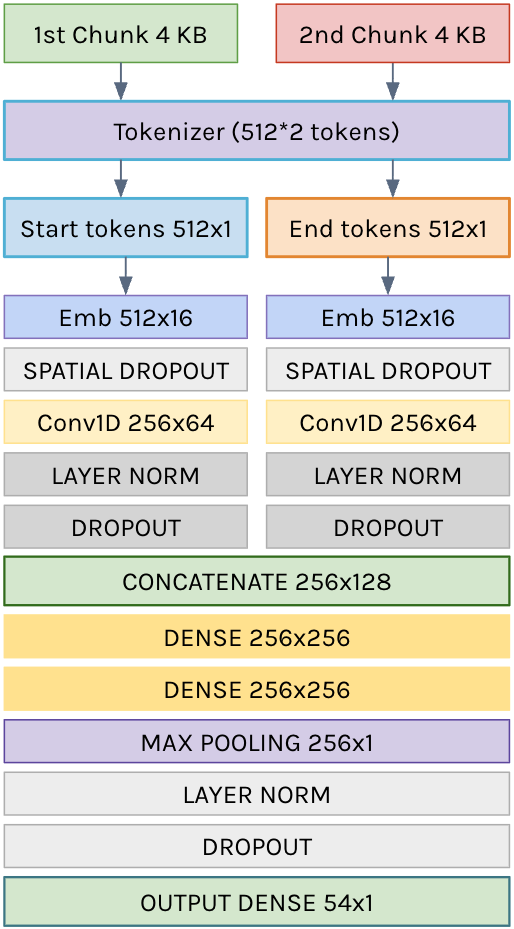}
    \caption{The dual-input convolutional neural network architecture of the \textsc{Labeler} model. The first three rows of blocks show how file contents are processed and the remaining blocks show the layers in the neural network model. The initial layers: embedding, spatial dropout, and 1-D Convolution, followed by layer normalization and dropout process, share the parameters, though they process the two token sequences separately.}
    \label{fig:model}
\end{figure}

The \textsc{Labeler} model is a dual-input convolutional neural network designed for programming language classification. The architecture processes both the beginning and end chunks of text content file types to capture syntactic patterns and structural information characteristic of different programming languages. The model architecture is shown in Figure \ref{fig:model}. As shown in the figure, several layers operate on each token sequence (start or end) independently. These layers have use weight sharing which enables efficient training of the model and a low memory footprint during inference.

The shared embedding layer maps discrete tokens to dense vector representations. A relatively low dimension of $d = 16$ was found to be sufficient to yield good results in this application, and desirable as we wish to minimize compute time and run on CPUs. The shared 1D-convolutional layer extracts local features from both sequences, using 64 filters with a kernel size of 4 and a stride of 2. This layer reduces the sequence length by half for each token sequence. The concatenation layer concatenates the two sequences along the feature dimension. The layers following concatenation are similar to the last few layers in Magika and represent the extraction of more detailed features before the final dense layer with softmax activation. All layers except the final dense layer use the rectified linear unit (ReLU)~\cite{Nair2010} activation function rather than its more complex variants~\cite{Clevert2016} to improve efficiency. The output of the last layer is the vector of probabilities that a given file belongs to each of the output file types. This architecture yields a model with approximately 715 KB of parameters that supports efficient training and inference.
Our proposed \textsc{Labeler} model is different from Magika in the following respects:
\begin{itemize}
    \item \textsc{Labeler} works on integer tokens computed by our custom tokenizer. The file contents are read as text (after decoding their UTF encoding), followed by conversion into fragments of text, each of which is then replaced by an integer ID. Magika does not use a tokenizer and operates directly on raw bytes.
    \item \textsc{Labeler} reads two chunks of files, while Magika reads three (start, middle and end). In all, \textsc{Labeler} processes up to 8KB of file contents while Magika is limited to 1.5 KB in the version described in \cite{fratantonio25_magika}.
    \item \textsc{Labeler} uses a different neural network to deal with the tokenized input, with the use of an embedding layer, which is different from Magika. In addition, several of the first few layers operate on each token sequence independently, and their outputs are finally concatenated to produce a single feature sequence for the file.
    \item Finally, \textsc{Labeler} is designed for text content file type detection, while Magika is designed for the detection of text content and binary file types.
\end{itemize}

\section{Evaluation} \label{sec:evaluation}
The \textsc{Labeler} model was evaluated against Magika as the baseline. As described in Section \ref{sec_related_work_model_based}, though there are other methods such as guesslang that also identify text content file types, Magika is reported to be the Pareto optimal solution today: significantly more accurate and has a much faster inference time, due to which we limited our experiments to comparison with Magika. All experiments used the Python library of Magika (version 1.0.1) installed from PyPI. Our custom tokenizer logic described in \ref{alg:tokenization} was implemented in Rust for efficiency. We note that because Magika calls out to underlying C libraries for byte reading and tensor calculations, and does not have any transformations to implement in Python, this is a fair comparison where each method will be leveraging Python libraries with compiled code for content extraction. 

For all file types, a probability threshold of 0.5 was used on the output of the \textsc{Labeler} to classify a given file as of a type. For files for which none of the file-type probabilities exceed this threshold, the predicted type is set to `unknown'. Magika also returns the label `unknown' when the predicted probability is below a threshold, though the threshold can vary by file type, as defined in the configuration.

\subsection{Datasets}
\subsubsection{Dataset Sources}

The training dataset is constructed from the BigCode \cite{bigcode} project, which provides a comprehensive collection of permissively licensed open-source file repositories. The primary source is The Stack dataset\cite{Kocetkov2022TheStack} (`The Stack dedup'), containing approximately 2.9 TB of deduplicated files across 358 file types with convenient streaming API access. For programming languages with insufficient representation in The Stack dataset- Verilog, MATLAB, ColdFusion, AppleScript, DM, and COBOL, the more comprehensive Stack V2 dataset \cite{lozhkov2024starcoder} was utilized (`The Stack v2 dedup'), encompassing 32 TB of deduplicated files across 658 file types. Only open-source files under permissive licenses (MIT, Apache 2.0, BSD-3-Clause, and BSD-2-Clause) were included, yielding approximately 9.75 million files, with an 80/10/10 split for training, validation, and testing, respectively.

\subsubsection{Language Coverage and Organization}

The final dataset encompasses \textit{fifty four} distinct programming languages spanning mainstream languages such as Python, JavaScript, and Java to specialized domain-specific languages including COBOL, VHDL, and Verilog. Language families are grouped in this model to improve classification accuracy: C and C++ are unified under a ``c-family'' label, Java and Groovy share a ``java-family'' classification, and JavaScript variants, including TypeScript, are grouped as ``js-family''. 
This hierarchical organization reduces the effective number of output classes while maintaining meaningful distinctions between programming paradigms. Since Magika does not use such language-family groupings, Magika's outputs are mapped to the corresponding labels used by the \textsc{Labeler} model, enabling comparison between the models. For example, if Magika predicts a file type of ``groovy'', it is first mapped to ``java-family'' before the evaluation of the metrics described in this study. The file types supported by the \textsc{Labeler} model are shown in Table
\ref{tab:stack_f1_comparison1}. File types ActionScript, AppleScript, ColdFusion, coldfusion and Java Server Pages are not supported by the latest version of Magika (version 1.0.1) at the time of this writing.

\subsection{Model Training}
The model was trained using the AdamW optimizer~\cite{Loshchilov2019} with learning rate $\alpha = 0.001$, $\beta_1 = 0.9$, $\beta_2 = 0.999$, and gradient clipping with $\text{clipnorm} = 1.0$. The loss function is sparse categorical cross-entropy, the standard loss for multi-class classification problems. Distributed training across the four GPUs on an EC2 instance (g4dn.12xlarge) with a batch size of 2048 and 40 training epochs takes less than eight hours to complete. All \textsc{Labeler} model-related implementation used the TensorFlow library, and the custom tokenizer for file content processing was done using a Rust implementation. The dropout rate for the first dropout layer (spatial dropout)~\cite{Tompson2015} was set as 0.1, while for the second it was 0.15, and for the third it was 0.2. All other layer parameters were set to their default values in the TensorFlow implementation.

\subsection{Inference Time Measurement}
Efficiency evaluation of both models was conducted using controlled timing experiments on a single CPU core with TensorFlow configured to use single-threaded execution and GPU acceleration disabled. This setup was intended to replicate its original use case: a file-identification tool that runs on client machines in the background with minimal impact on other processes. As mentioned earlier, a model that requires multiple CPUs for inference can lead to undesirable user experience and our measurement method considers such applications also.

\begin{figure}
    \centering
    \begin{tikzpicture}
\begin{axis}[
  callisto axis,
  title={Inference Time},
  xlabel={Inference time (ms)},
  ylabel={Number of inference calls},
  xmin=0, xmax=5,
  ymin=0, ymax=76000,
  xtick={0,1,2,3,4,5},
  ytick={0,10000,20000,30000,40000,50000,60000,70000},
  scaled y ticks=false,
  yticklabel={\pgfmathprintnumber[fixed,precision=0]{\tick}},
  legend pos=north east
]
\addplot+[ybar, bar width=2pt, callisto labeler fill, forget plot] coordinates {
  (0.975,1467) (1.025,72596) (1.075,12907) (1.125,5403) (1.175,3096) (1.225,1860) (1.275,975) (1.325,533) (1.375,413) (1.425,276) (1.475,157) (1.525,80) (1.575,38) (1.625,16) (1.675,12) (1.725,5) (1.775,1) (1.825,2) (1.875,4) (1.925,2) (1.975,3) (2.025,2) (2.075,2) (2.125,1) (2.225,2) (2.275,2) (2.325,1) (2.675,1) (2.775,1) (2.825,2) (2.875,1) (2.975,1) (3.175,3) (3.525,1) (3.625,1) (3.725,2) (3.775,1) (3.875,1) (3.975,1) (4.025,1) (4.175,1) (4.225,1) (4.375,1) (4.475,1) (4.575,1) (4.725,1) (4.775,1) (4.825,1)
};
\addplot+[ybar, bar width=2pt, callisto magika fill, forget plot] coordinates {
  (3.675,967) (3.725,14632) (3.775,23175) (3.825,25932) (3.875,18873) (3.925,9001) (3.975,3495) (4.025,1449) (4.075,642) (4.125,355) (4.175,225) (4.225,181) (4.275,122) (4.325,109) (4.375,73) (4.425,76) (4.475,47) (4.525,33) (4.575,28) (4.625,29) (4.675,23) (4.725,20) (4.775,19) (4.825,14) (4.875,13) (4.925,23)
};
\addlegendimage{area legend,draw=white,fill=CallistoBlue,fill opacity=0.55}
\addlegendentry{Labeler (single CPU)}
\addlegendimage{area legend,draw=white,fill=CallistoGreen,fill opacity=0.55}
\addlegendentry{Magika (no CPU limits)}
\end{axis}
\end{tikzpicture}
    \caption{Inference time benchmarked by setting \textsc{Labeler} to use one CPU core only, without any such limitation on Magika. To measure \textsc{Labeler} inference times, a random byte array is generated of 8 Kb size and fed to a Rust implementation of the custom tokenizer the output of which is fed to the neural network model. Similarly, to measure Magika's inference times, a random byte array of size 1.5 Kb is generated and passed as input to the Python interface of Magika. The average time taken for hundred thousand such calls is shown here}
    \label{fig:inf_times}
\end{figure}
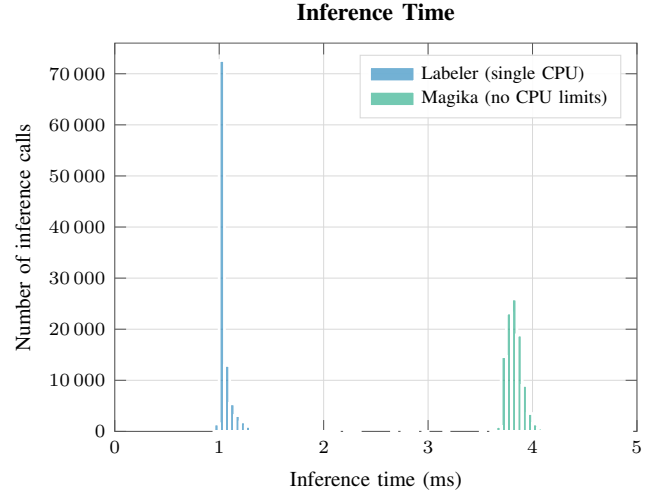

To measure the \textsc{Labeler}'s inference speed, a hundred thousand random byte sequences of 8,192 bytes each were generated. Time taken to tokenize each byte sequence using a rust implementation of the tokenizer and the neural network model to generate predictions on the tokens is measured. Similarly, for Magika, a hundred thousand random byte sequences of 1,536 bytes are generated. Inference time is measured through Magika's \texttt{identify\_bytes} method. Both functions exclude the first 100 iterations to account for model initialization overhead and report average inference times in milliseconds per file. Note that benchmarking was performed on an Apple MacBook Pro (M1 Max) with \textsc{Labeler} limited to one CPU core. This test was designed to minimize any latency concerns in passing the file contents as arguments to the Magika library.  The controlled experimental setup ensures fair comparison between the two approaches, with the results displayed in Figure \ref{fig:inf_times}. \textsc{Labeler} achieves approximately 1.05 ms per file compared to Magika's 3.85 ms per file, demonstrating nearly 3.7$\times$ faster inference speed while also processing $5.3\times$ more bytes, for a net $19.6\times$ advantage per-byte. 

\subsection{Accuracy Results}

We evaluate model performance using standard classification metrics. Precision measures the proportion of correct positive predictions for a file type, and Recall measures the proportion of actual positives correctly identified. F1-score is the harmonic mean of Precision and Recall. We report both Macro F1 (unweighted average of per-class F1-scores) and Weighted F1 (average of per-class F1-scores weighted by class proportion) across all classes. Any file classified as `unknown' due to its predicted probability not meeting the threshold of 0.5 is counted as a false negative for that file's type.

\begin{table}[!h]
\centering
\caption{Summary of performance metrics comparing \textsc{Labeler} and Magika. \textsc{Labeler} demonstrates superior accuracy with both macro and weighted F1-scores. Most significantly, \textsc{Labeler} has 3.7x faster inference speed even when it is limited to use a single CPU core, unlike Magika.}
\label{tab:performance_summary}
\begin{tabular}{@{}l|c|c@{}}
\textbf{Metrics} & \textbf{\textsc{Labeler}} & \textbf{Magika} \\
Macro F1 & 0.9804 & 0.9057 \\
Weighted F1 & 0.982 & 0.921 \\
Inference time per file & 1.05 ms & 3.85 ms \\
Model parameters size & 715 Kb & 1 Mb\\
\end{tabular}
\end{table}

A summary of the results of the comparison of \textsc{Labeler} with Magika is shown in Table \ref{tab:performance_summary}, showing a superior performance in terms of Macro-F1 and Weighted-F1. The $Macro F_1$ score of \textsc{Labeler} is 0.98, which is significantly higher compared to Magika's 0.91. Similarly, the $Weighted F_1$ score of \textsc{Labeler} is 0.982, which is significantly higher than Magika's 0.92. The most significant result is likely the reduction in inference time per file, with \textsc{Labeler} being about 3.7 times faster than Magika. The faster inference is likely the result of using a smaller model (\textsc{Labeler} is $28\%$ smaller than Magika) and the neural network architecture that processes each file chunk independently in the first few layers. An estimate of the size of the custom tokenizer implementation in Rust can be obtained from the compiled Python wheel (220 KB), which includes the vocabulary definition. Hence, the model and tokenizer together have a lower memory footprint than Magika.

Table \ref{tab:stack_f1_comparison1} shows the per-class F1-score for each of the supported fifty-four file types. Across all file types, \textsc{Labeler} achieves a higher F1-score than Magika, with the difference being significantly greater in some cases. 

\begin{table}[!h]
\centering
\caption{Performance comparison for some of the supported file types. Bold values indicate the best results. -- indicate unsupported types.}
\label{tab:stack_f1_comparison1}
\begin{tabular}{@{}l|r|r|c|c@{}}
\textbf{Language}  & \textbf{Train Size} & \textbf{Test Size} & \multicolumn{2}{c}{\textbf{F1-Score}} \\
& & & \textbf{Labeler} & \textbf{Magika} \\
\hline
actionscript & 103768 & 11497 & \textbf{0.9895} & - \\
applescript & 11214 & 1246 & \textbf{0.9730} & - \\
asp & 65067 & 7259 & \textbf{0.9881} & 0.4893 \\
assembly & 178619 & 19981 & \textbf{0.9896} & 0.9171 \\
batchfile & 188072 & 20967 & \textbf{0.9565} & 0.9207 \\
c\# & 201939 & 22431 & \textbf{0.9959} & 0.9632 \\
c-family & 367257 & 40670 & \textbf{0.9881} & 0.9670 \\
clojure & 102617 & 11386 & \textbf{0.9956} & 0.9826 \\
cmake & 147208 & 16316 & \textbf{0.9800} & 0.9710 \\
cobol & 19392 & 2154 & \textbf{0.9627} & 0.7297 \\
coffeescript & 185240 & 20612 & \textbf{0.9881} & 0.9351 \\
coldfusion & 70256 & 7806 & \textbf{0.9628} & - \\
common-lisp & 77710 & 8605 & \textbf{0.9892} & 0.9427 \\
css & 199456 & 22124 & \textbf{0.9967} & 0.9366 \\
csv & 192029 & 21291 & \textbf{0.9416} & 0.8364 \\
dart & 207456 & 23080 & \textbf{0.9960} & 0.9873 \\
dm & 17999 & 1999 & \textbf{0.9505} & 0.8159 \\
dockerfile & 199626 & 22227 & \textbf{0.9973} & 0.9972 \\
elixir & 205714 & 22843 & \textbf{0.9963} & 0.9896 \\
erlang & 83212 & 9236 & \textbf{0.9978} & 0.9923 \\
fortran & 123033 & 13755 & \textbf{0.9751} & 0.8458 \\
go & 207841 & 23095 & \textbf{0.9989} & 0.9931 \\
haskell & 206448 & 22938 & \textbf{0.9974} & 0.9911 \\
html & 180727 & 20179 & \textbf{0.9440} & 0.6326 \\
java-family & 409083 & 45492 & \textbf{0.9941} & 0.9260 \\
java-server-pages & 178188 & 19837 & \textbf{0.9822} & - \\
javascript-family & 407378 & 45292 & \textbf{0.9917} & 0.9645 \\
json & 174040 & 19364 & \textbf{0.9820} & 0.9310 \\
julia & 210525 & 23428 & \textbf{0.9924} & 0.9696 \\
kotlin & 203386 & 22604 & \textbf{0.9956} & 0.9696 \\
makefile & 180917 & 20129 & \textbf{0.9921} & 0.9700 \\
matlab & 52516 & 5999 & \textbf{0.9902} & 0.9659 \\
objective-c++ & 46293 & 5154 & \textbf{0.9577} & 0.7579 \\
ocaml & 119684 & 13288 & \textbf{0.9942} & 0.9591 \\
pascal & 91491 & 10153 & \textbf{0.9953} & 0.8586 \\
perl & 139063 & 15407 & \textbf{0.9709} & 0.8141 \\
php & 208007 & 23096 & \textbf{0.9819} & 0.9444 \\
powershell & 204827 & 22788 & \textbf{0.9810} & 0.9632 \\
python & 201741 & 22424 & \textbf{0.9887} & 0.9673 \\
r & 30548 & 3398 & \textbf{0.9578} & 0.8578 \\
ruby & 199115 & 22128 & \textbf{0.9932} & 0.9398 \\
rust & 194748 & 21654 & \textbf{0.9982} & 0.9895 \\
scala & 205003 & 22817 & \textbf{0.9954} & 0.9743 \\
shell & 197274 & 21926 & \textbf{0.9646} & 0.9378 \\
sql & 191799 & 21299 & \textbf{0.9777} & 0.9621 \\
swift & 208704 & 23242 & \textbf{0.9979} & 0.9900 \\
tex & 169920 & 18868 & \textbf{0.9893} & 0.8482 \\
text & 60000 & 21540 & \textbf{0.7903} & 0.3132 \\
toml & 192270 & 21390 & \textbf{0.9927} & 0.9700 \\
verilog & 52474 & 5999 & \textbf{0.9952} & 0.9776 \\
vhdl & 46265 & 5169 & \textbf{0.9961} & 0.9669 \\
visual-basic & 114585 & 12735 & \textbf{0.9837} & 0.9292 \\
xml & 185454 & 20438 & \textbf{0.9867} & 0.8892 \\
yaml & 182824 & 20440 & \textbf{0.9814} & 0.9421 \\
\end{tabular}
\end{table}

\textsc{Labeler}'s dominance over Magica is a reflection of careful design in model and feature processing, and reflects a challenge in the canonical deep-learning paradigm. While a deep model can often perform better given enough training data and compute (e.g., Convolutional Neural Networks displacing classic Computer Vision designs, Recurrent Neural Networks displacing much of stemming and classical Natural Language Processing, and now Transformers yet-still displacing RNNs and many NLP sub-tasks like entity recognition), this advantage is not necessarily free from a computational efficiency perspective. Even a simpler tokenizer as we use in Algorithm \ref{alg:tokenization} carefully designed for the domain has yielded significant advantage in accuracy for a tight compute budget. Such runtime constraints are common in cybersecurity applications~\cite{Raff2020a}. As mentioned earlier, if the predicted probability for a file is less than the threshold of 0.5 for any type, then the file type is taken to be `unknown' and it is counted as an FP. For all the files in the test set, only $0.8\%$ of the files have a predicted file type of `unknown'.

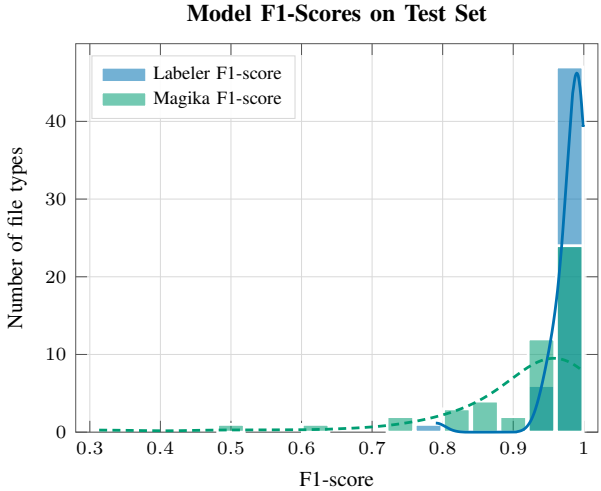
\begin{figure}[!h]
    \centering
    \begin{tikzpicture}
\begin{axis}[
  callisto axis,
  title={Model F1-Scores on Test Set},
  xlabel={F1-score},
  ylabel={Number of file types},
  xmin=0.28, xmax=1.02,
  ymin=0, ymax=50,
  xtick={0.3,0.4,0.5,0.6,0.7,0.8,0.9,1.0},
  ytick={0,10,20,30,40},
  legend pos=north west
]
\addplot+[ybar, bar width=10pt, callisto labeler fill, forget plot] coordinates {
  (0.78,1) (0.94,6) (0.98,47)
};
\addplot+[ybar, bar width=10pt, callisto magika fill, forget plot] coordinates {
  (0.5,1) (0.62,1) (0.74,2) (0.82,3) (0.86,4) (0.9,2) (0.94,12) (0.98,24)
};
\addplot+[color=CallistoBlue, smooth, no marks, forget plot] coordinates {
  (0.7903,1.170692) (0.793445,1.139948) (0.797638,1.012787) (0.802879,0.764755) (0.811265,0.358736) (0.816506,0.184433) (0.821747,0.081788) (0.829085,0.020437) (0.841664,0.000966) (0.899317,0.016454) (0.905607,0.066309) (0.909799,0.150692) (0.912944,0.263909) (0.916089,0.441344) (0.919234,0.705754) (0.922378,1.080878) (0.925523,1.588355) (0.928668,2.244275) (0.931813,3.056195) (0.936006,4.375566) (0.941247,6.357409) (0.947536,9.112417) (0.952777,11.706312) (0.95697,14.092556) (0.961163,16.961124) (0.964308,19.574216) (0.968501,23.825117) (0.972694,28.946513) (0.983176,42.379361) (0.985273,44.255071) (0.987369,45.550071) (0.988418,45.945383) (0.989466,46.158134) (0.990514,46.180507) (0.991562,46.00739) (0.992611,45.636552) (0.994707,44.307756) (0.996804,42.235965) (0.9989,39.507844) (0.9989,39.507844)
};
\addplot+[color=CallistoGreen, smooth, densely dashed, no marks, forget plot] coordinates {
  (0.3132,0.26734) (0.347572,0.238237) (0.406004,0.175647) (0.443813,0.221444) (0.495371,0.283217) (0.605361,0.316267) (0.656919,0.455826) (0.687853,0.597012) (0.711914,0.771532) (0.732537,0.983398) (0.75316,1.264324) (0.770346,1.558702) (0.787532,1.915313) (0.804718,2.345074) (0.818466,2.753999) (0.832215,3.236252) (0.842527,3.656389) (0.852838,4.13409) (0.86315,4.673334) (0.876898,5.485274) (0.894084,6.611765) (0.914708,7.97222) (0.925019,8.567712) (0.931893,8.903473) (0.938768,9.175351) (0.945642,9.371942) (0.952517,9.483815) (0.959391,9.504113) (0.966265,9.429016) (0.97314,9.258036) (0.980014,8.994102) (0.986888,8.643449) (0.993763,8.215293) (0.9972,7.975732) (0.9972,7.975732)
};
\addlegendimage{area legend,draw=white,fill=CallistoBlue,fill opacity=0.55}
\addlegendentry{Labeler F1-score}
\addlegendimage{area legend,draw=white,fill=CallistoGreen,fill opacity=0.55}
\addlegendentry{Magika F1-score}
\end{axis}
\end{tikzpicture}
    \caption{Distribution of F1 score for the different file types. As seen, \textsc{Labeler} has a much narrower distribution than Magika. Both model gave least F1-scores for ``text'' file type. However, for other types, the results for \textsc{Labeler} show an F1-Score of at least 0.94.}
    \label{fig:f1s}
\end{figure}

\begin{figure}[!h]
    \centering
    \begin{tikzpicture}
\begin{axis}[
  callisto axis,
  title={Impact of Training Set Size},
  xlabel={Number of files in training set},
  ylabel={Classification metric},
  xmin=-0.25, xmax=5.25,
  ymin=0.889, ymax=0.991,
  xtick={0,1,2,3,4,5},
  xticklabels={2\,000,5\,000,10\,000,20\,000,50\,000,All files},
  x tick label style={rotate=35,anchor=east,font=\scriptsize},
  ytick={0.90,0.92,0.94,0.96,0.98},
  legend style={at={(0.02,0.98)},anchor=north west}
]
\addplot+[callisto labeler line] coordinates {(0,0.9239) (1,0.937913) (2,0.942802) (3,0.951874) (4,0.958439) (5,0.986772)};\addlegendentry{Labeler precision}
\addplot+[color=CallistoGreen, no marks] coordinates {(0,0.9362) (1,0.9362) (2,0.9362) (3,0.9362) (4,0.9362) (5,0.9362)};\addlegendentry{Magika precision}
\addplot+[callisto labeler line, densely dashed, mark=square*] coordinates {(0,0.934491) (1,0.948339) (2,0.954193) (3,0.958513) (4,0.962469) (5,0.974952)};\addlegendentry{Labeler recall}
\addplot+[callisto magika line, no marks] coordinates {(0,0.893424) (1,0.893424) (2,0.893424) (3,0.893424) (4,0.893424) (5,0.893424)};\addlegendentry{Magika recall}
\end{axis}
\end{tikzpicture}
    \caption{Improvement in precision and recall of \textsc{Labeler} with increase in training set size. Even with a very low number of files per type in the training set, \textsc{Labeler} quickly produces a higher accuracy than Magika using all available training data. This shows our method has excellent sample efficiency for rare and future new file types.}
    \label{fig:train_set_size}
\end{figure}
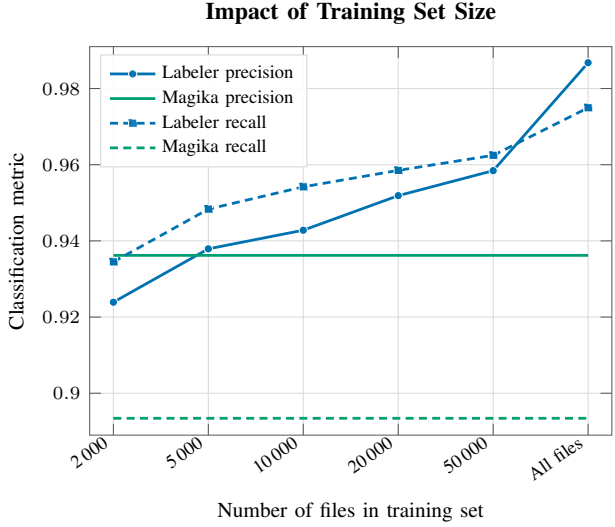

To measure the impact of training set size on the classification performance of \textsc{Labeler}, we trained the model by setting an upper limit on the number of files in each file type. The resulting precision and recall are shown in Figure \ref{fig:train_set_size}. Note that the same test set files as in the results described above are used here. It can be seen that \textsc{Labeler} 's recall is higher than Magika's, even with only 2000 training files per file type. The precision is lower than Magika in the case when the training set size is 2000, but higher in all other cases. These results indicate that the \textsc{Labeler} architecture is able to improve upon the state-of-the-art model even with a very low number of training set files. This again shows the value in having a customized tokenizer and architecture, allowing a large improvement in sample efficiency. Such advantage is critical from an operational maintenance perspective, as new file types emerge with intrinsically limited availability, we are more likely to be able to produce a satisfying production model in a timely fashion. 

The key insights from the evaluation are listed below:
\begin{itemize}
    \item \textsc{Labeler} gives a relatively consistent performance with an F1-score higher than 0.94 for all types except text (.txt files).
    \item Both \textsc{Labeler} and Magika were measured to have the worst performance for text (.txt files). This is not very surprising since these are text files from GitHub projects, that are found to have a mix of code/configuration type files in them. For examples, a few text files have CMake commands in them as part of instructions.
    \item Even for file types that have fewer samples in our training dataset, the \textsc{Labeler} has a higher F1-score than Magika. For the five types supported by both models with the least training set size - DM, COBOL, R, VHDL, Objective-C++, \textsc{Labeler} has a $Macro F_1$ of 0.96 compared to Magik's 0.83. The model architecture seems to be able to learn enough details from even comparatively small training sets for these languages, so it is reasonable to expect the model to learn from other such rare or new languages as and when training data becomes available for them.
\end{itemize}

\subsubsection{Customizability of \textsc{Labeler}}
\begin{figure}[H]
    \centering
    \begin{tikzpicture}
\begin{axis}[
  callisto axis,
  title={Impact of Training Set Size on Rare Types},
  xlabel={Training set size per type},
  ylabel={Classification metric},
  xmin=-0.25, xmax=5.25,
  ymin=0.70, ymax=1.005,
  xtick={0,1,2,3,4,5},
  xticklabels={2\,000,5\,000,10\,000,20\,000,50\,000,All files},
  x tick label style={rotate=35,anchor=east,font=\scriptsize},
  ytick={0.70,0.75,0.80,0.85,0.90,0.95,1.00},
  legend style={at={(0.02,0.52)},anchor=west}
]
\addplot+[callisto magika line, no marks] coordinates {(0,0.7174) (1,0.7174) (2,0.7174) (3,0.7174) (4,0.7174) (5,0.7174)};\addlegendentry{Magika recall}
\addplot+[callisto labeler line] coordinates {(0,0.73344) (1,0.79264) (2,0.80124) (3,0.8335) (4,0.8736) (5,0.98932)};\addlegendentry{Labeler precision}
\addplot+[color=CallistoGreen, no marks] coordinates {(0,0.99004) (1,0.99004) (2,0.99004) (3,0.99004) (4,0.99004) (5,0.99004)};\addlegendentry{Magika precision}
\addplot+[callisto labeler line, densely dashed, mark=square*] coordinates {(0,0.93116) (1,0.94502) (2,0.95474) (3,0.96016) (4,0.95638) (5,0.94208)};\addlegendentry{Labeler recall}
\end{axis}
\end{tikzpicture}
    \caption{Relation of precision and recall for the rarest five file types (DM, COBOL, R, VHDL, and Objective-C++) with increase in training set size. The drop in recall beyond a size of 50000 is likely due to the fact that all these types have less than 50000 files in the original training set.}
    \label{fig:train_set_size2}
\end{figure}
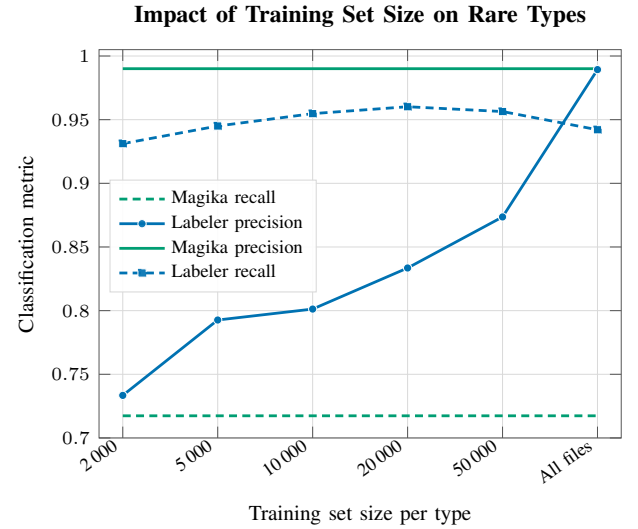

The \textsc{Labeler} model is relatively easy to customize. Given the availability of ample open-source code files and the empirically demonstrated ability of the model to identify file types, even when there are comparatively few files of those types in the training set, adding support for new file types is straightforward. An interesting result on the impact of varying the training set size on the detection of the rare file types (DM, COBOL, R, VHDL, and Objective-C++) is shown in Figure \ref{fig:train_set_size2}. As expected, the average precision increases with an increase in train set size. The average recall also increases initially, but then drops slightly, specifically when the maximum training set size per type is 50000 or higher. This is an expected result due to the fact that all these types have a training set size of less than 50000 in the original training set, leading to an imbalance in the training set when larger sizes. Note that the same test set as in other experiment results is also used here. Future research could explore oversampling files from under-represented types or augmenting data through two approaches: (1) combining portions of existing files using language parsers, or (2) generating synthetic files using LLMs.

\section{Conclusion} \label{sec:conclusion}
In this paper, we introduced a lightweight neural network model, Labeler, designed to efficiently and accurately identify text content file types from their contents. A custom tokenizer algorithm that can be efficiently implemented is also presented, which can process text content files into a format suitable for the model to process. Similar to the method in Magika, the model reads only chunks of a given file to predict its file type. Based on extensive experiments on open source files, it was found that \textsc{Labeler} has an Macro $F1$ that is $8\%$ higher than Magika while being 3.7 times faster for inference and about $28\%$ smaller in terms of model parameter size. These results indicate that the newly proposed \textsc{Labeler} model has superior performance for the identification of text content file types.

\FloatBarrier 

\bibliographystyle{IEEEtran}
\bibliography{references}

\begin{thebibliography}{10}
\providecommand{\url}[1]{#1}
\csname url@samestyle\endcsname
\providecommand{\newblock}{\relax}
\providecommand{\bibinfo}[2]{#2}
\providecommand{\BIBentrySTDinterwordspacing}{\spaceskip=0pt\relax}
\providecommand{\BIBentryALTinterwordstretchfactor}{4}
\providecommand{\BIBentryALTinterwordspacing}{\spaceskip=\fontdimen2\font plus
\BIBentryALTinterwordstretchfactor\fontdimen3\font minus \fontdimen4\font\relax}
\providecommand{\BIBforeignlanguage}[2]{{%
\expandafter\ifx\csname l@#1\endcsname\relax
\typeout{** WARNING: IEEEtran.bst: No hyphenation pattern has been}%
\typeout{** loaded for the language `#1'. Using the pattern for}%
\typeout{** the default language instead.}%
\else
\language=\csname l@#1\endcsname
\fi
#2}}
\providecommand{\BIBdecl}{\relax}
\BIBdecl

\bibitem{fratantonio25_magika}
Y.~Fratantonio, L.~Invernizzi, L.~Farah, K.~Thomas, M.~Zhang, A.~Albertini, F.~Galilee, G.~Metitieri, J.~Cretin, A.~Petit-Bianco, D.~Tao, and E.~Bursztein, ``{Magika: AI-Powered Content-Type Detection},'' in \emph{Proceedings of the International Conference on Software Engineering (ICSE)}, April 2025.

\bibitem{file_nix}
\BIBentryALTinterwordspacing
``File — a file type guesser,'' 2026. [Online]. Available: \url{https://www.darwinsys.com/file/}
\BIBentrySTDinterwordspacing

\bibitem{libmagic}
\BIBentryALTinterwordspacing
``libmagic(3) - linux manual page,'' 2023, describes the library functions for file type identification. [Online]. Available: \url{https://man7.org/linux/man-pages/man3/libmagic.3.html}
\BIBentrySTDinterwordspacing

\bibitem{ApacheTika}
\BIBentryALTinterwordspacing
{Apache Software Foundation}, ``{Apache Tika} toolkit for content analysis,'' 2025. [Online]. Available: \url{https://tika.apache.org/}
\BIBentrySTDinterwordspacing

\bibitem{exif}
\BIBentryALTinterwordspacing
``Exiftool,'' 2026. [Online]. Available: \url{https://exiftool.org/}
\BIBentrySTDinterwordspacing

\bibitem{trid}
\BIBentryALTinterwordspacing
``Trid,'' 2025. [Online]. Available: \url{https://mark0.net/soft-trid-e.html}
\BIBentrySTDinterwordspacing

\bibitem{guesslang}
\BIBentryALTinterwordspacing
``guesslang: tool to detect programming language of a given source,'' 2021. [Online]. Available: \url{https://github.com/yoeo/guesslang}
\BIBentrySTDinterwordspacing

\bibitem{Cheng_16}
\BIBentryALTinterwordspacing
H.-T. Cheng, L.~Koc, J.~Harmsen, T.~Shaked, T.~Chandra, H.~Aradhye, G.~Anderson, G.~Corrado, W.~Chai, M.~Ispir, R.~Anil, Z.~Haque, L.~Hong, V.~Jain, X.~Liu, and H.~Shah, ``Wide \& deep learning for recommender systems,'' in \emph{Proceedings of the 1st Workshop on Deep Learning for Recommender Systems}, ser. DLRS 2016.\hskip 1em plus 0.5em minus 0.4em\relax New York, NY, USA: Association for Computing Machinery, 2016, p. 7–10. [Online]. Available: \url{https://doi.org/10.1145/2988450.2988454}
\BIBentrySTDinterwordspacing

\bibitem{guesslang_VS_code}
\BIBentryALTinterwordspacing
``Visual studio code september 2021 (version 1.60) - automatic language detection,'' 2021. [Online]. Available: \url{https://code.visualstudio.com/updates/v1_60#_automatic-language-detection}
\BIBentrySTDinterwordspacing

\bibitem{vscode_guesslang}
\BIBentryALTinterwordspacing
``Vscode language detection: nodejs cli around guesslang,'' 2026. [Online]. Available: \url{https://github.com/microsoft/vscode-languagedetection}
\BIBentrySTDinterwordspacing

\bibitem{tika_magika}
\BIBentryALTinterwordspacing
``Magikadetector: Apache tika interface documentation,'' 2025. [Online]. Available: \url{https://tika.apache.org/3.1.0/api/org/apache/tika/detect/magika/MagikaDetector.html}
\BIBentrySTDinterwordspacing

\bibitem{FITZGERALD2012S44}
\BIBentryALTinterwordspacing
S.~Fitzgerald, G.~Mathews, C.~Morris, and O.~Zhulyn, ``Using nlp techniques for file fragment classification,'' \emph{Digital Investigation}, vol.~9, pp. S44--S49, 2012, the Proceedings of the Twelfth Annual DFRWS Conference. [Online]. Available: \url{https://www.sciencedirect.com/science/article/pii/S1742287612000333}
\BIBentrySTDinterwordspacing

\bibitem{Wang_18}
F.~Wang, T.-T. Quach, J.~Wheeler, J.~B. Aimone, and C.~D. James, ``Sparse coding for n-gram feature extraction and training for file fragment classification,'' \emph{IEEE Transactions on Information Forensics and Security}, vol.~13, no.~10, pp. 2553--2562, 2018.

\bibitem{Mittal_21}
G.~Mittal, P.~Korus, and N.~Memon, ``Fifty: Large-scale file fragment type identification using convolutional neural networks,'' \emph{IEEE Transactions on Information Forensics and Security}, vol.~16, pp. 28--41, 2021.

\bibitem{Kristian_23}
K.~Skračić, J.~Petrović, and P.~Pale, ``Bytercnn: Enhancing file fragment type identification with recurrent and convolutional neural networks,'' \emph{IEEE Access}, vol.~11, pp. 138\,176--138\,187, 2023.

\bibitem{McDaniel:2003:CBF:820756.821828}
\BIBentryALTinterwordspacing
M.~McDaniel and M.~H. Heydari, ``Content {Based} {File} {Type} {Detection} {Algorithms},'' in \emph{Proceedings of the 36th {Annual} {Hawaii} {International} {Conference} on {System} {Sciences} ({HICSS}'03) - {Track} 9 - {Volume} 9}.\hskip 1em plus 0.5em minus 0.4em\relax Washington, DC, USA: IEEE Computer Society, 2003, p. 332.1, series Title: HICSS '03. [Online]. Available: \url{http://dl.acm.org/citation.cfm?id=820756.821828}
\BIBentrySTDinterwordspacing

\bibitem{Pal2009}
\BIBentryALTinterwordspacing
A.~Pal and N.~Memon, ``The {Evolution} of {File} {Carving},'' \emph{IEEE Signal Processing Magazine}, vol.~26, no.~2, pp. 59--71, Mar. 2009. [Online]. Available: \url{http://ieeexplore.ieee.org/document/4806206/}
\BIBentrySTDinterwordspacing

\bibitem{Roussev:2009:FFC:1683311.1684973}
\BIBentryALTinterwordspacing
V.~Roussev and S.~L. Garfinkel, ``File {Fragment} {Classification}-{The} {Case} for {Specialized} {Approaches},'' in \emph{Proceedings of the 2009 {Fourth} {International} {IEEE} {Workshop} on {Systematic} {Approaches} to {Digital} {Forensic} {Engineering}}.\hskip 1em plus 0.5em minus 0.4em\relax Washington, DC, USA: IEEE Computer Society, 2009, pp. 3--14, series Title: SADFE '09. [Online]. Available: \url{http://dx.doi.org/10.1109/SADFE.2009.21}
\BIBentrySTDinterwordspacing

\bibitem{Axelsson2010S24}
\BIBentryALTinterwordspacing
S.~Axelsson, ``The {Normalised} {Compression} {Distance} as a file fragment classifier,'' \emph{Digital Investigation}, vol. Volume 7,, pp. S24 -- S31, 2010. [Online]. Available: \url{http://www.sciencedirect.com/science/article/pii/S1742287610000319}
\BIBentrySTDinterwordspacing

\bibitem{Gopal2011}
\BIBentryALTinterwordspacing
S.~Gopal, Y.~Yang, K.~Salomatin, and J.~Carbonell, ``Statistical {Learning} for {File}-{Type} {Identification},'' in \emph{2011 10th {International} {Conference} on {Machine} {Learning} and {Applications} and {Workshops}}, vol.~1.\hskip 1em plus 0.5em minus 0.4em\relax IEEE, Dec. 2011, pp. 68--73, issue: DiiD. [Online]. Available: \url{http://ieeexplore.ieee.org/document/6146945/}
\BIBentrySTDinterwordspacing

\bibitem{Poisel2013}
\BIBentryALTinterwordspacing
R.~Poisel and S.~Tjoa, ``A {Comprehensive} {Literature} {Review} of {File} {Carving},'' in \emph{2013 {International} {Conference} on {Availability}, {Reliability} and {Security}}.\hskip 1em plus 0.5em minus 0.4em\relax IEEE, Sep. 2013, pp. 475--484. [Online]. Available: \url{http://ieeexplore.ieee.org/document/6657278/}
\BIBentrySTDinterwordspacing

\bibitem{raff_lzjd_digest}
\BIBentryALTinterwordspacing
E.~Raff and C.~K. Nicholas, ``Lempel-{Ziv} {Jaccard} {Distance}, an effective alternative to ssdeep and sdhash,'' \emph{Digital Investigation}, Feb. 2018, arXiv: 1708.03346. [Online]. Available: \url{https://doi.org/10.1016/j.diin.2017.12.004}
\BIBentrySTDinterwordspacing

\bibitem{chang_fbhash_2019}
\BIBentryALTinterwordspacing
D.~Chang, M.~Ghosh, S.~K. Sanadhya, M.~Singh, and D.~R. White, ``{FbHash}: {A} {New} {Similarity} {Hashing} {Scheme} for {Digital} {Forensics},'' \emph{Digital Investigation}, vol.~29, pp. S113--S123, Jul. 2019. [Online]. Available: \url{https://www.sciencedirect.com/science/article/pii/S1742287619301550}
\BIBentrySTDinterwordspacing

\bibitem{raff_lzjd_2017}
\BIBentryALTinterwordspacing
E.~Raff and C.~Nicholas, ``An {Alternative} to {NCD} for {Large} {Sequences}, {Lempel}-{Ziv} {Jaccard} {Distance},'' in \emph{Proceedings of the 23rd {ACM} {SIGKDD} {International} {Conference} on {Knowledge} {Discovery} and {Data} {Mining} - {KDD} '17}.\hskip 1em plus 0.5em minus 0.4em\relax New York, New York, USA: ACM Press, 2017, pp. 1007--1015. [Online]. Available: \url{http://dl.acm.org/citation.cfm?doid=3097983.3098111}
\BIBentrySTDinterwordspacing

\bibitem{Clemens2015}
\BIBentryALTinterwordspacing
J.~Clemens, ``Automatic classification of object code using machine learning,'' \emph{Digital Investigation}, vol.~14, pp. S156--S162, 2015. [Online]. Available: \url{http://www.sciencedirect.com/science/article/pii/S1742287615000523}
\BIBentrySTDinterwordspacing

\bibitem{Hand2012}
\BIBentryALTinterwordspacing
S.~Hand, Z.~Lin, G.~Gu, and B.~Thuraisingham, ``Bin-{Carver}: {Automatic} recovery of binary executable files,'' \emph{Digital Investigation}, vol.~9, pp. S108--S117, 2012. [Online]. Available: \url{http://www.sciencedirect.com/science/article/pii/S1742287612000394}
\BIBentrySTDinterwordspacing

\bibitem{vandenBos:2011:BDL:1985793.1985887}
\BIBentryALTinterwordspacing
J.~van~den Bos and T.~van~der Storm, ``Bringing {Domain}-specific {Languages} to {Digital} {Forensics},'' in \emph{Proceedings of the 33rd {International} {Conference} on {Software} {Engineering}}.\hskip 1em plus 0.5em minus 0.4em\relax New York, NY, USA: ACM, 2011, pp. 671--680, series Title: ICSE '11. [Online]. Available: \url{http://doi.acm.org/10.1145/1985793.1985887}
\BIBentrySTDinterwordspacing

\bibitem{Wilgenbus2013}
\BIBentryALTinterwordspacing
E.~F. Wilgenbus, ``The file fragment classification problem : {A} combined neural network and linear programming discriminant model approach,'' Ph.D. dissertation, North-West University, 2013, issue: April. [Online]. Available: \url{https://repository.nwu.ac.za/handle/10394/10215}
\BIBentrySTDinterwordspacing

\bibitem{Raff2020b}
\BIBentryALTinterwordspacing
E.~Raff, W.~Fleshman, R.~Zak, H.~S. Anderson, B.~Filar, and M.~McLean, ``Classifying {Sequences} of {Extreme} {Length} with {Constant} {Memory} {Applied} to {Malware} {Detection},'' in \emph{The {Thirty}-{Fifth} {AAAI} {Conference} on {Artificial} {Intelligence}}, 2021, arXiv: 2012.09390. [Online]. Available: \url{http://arxiv.org/abs/2012.09390}
\BIBentrySTDinterwordspacing

\bibitem{MalConv}
\BIBentryALTinterwordspacing
E.~Raff, J.~Barker, J.~Sylvester, R.~Brandon, B.~Catanzaro, and C.~Nicholas, ``{Malware Detection by Eating a Whole EXE},'' in \emph{AAAI Workshop on Artificial Intelligence for Cyber Security}, oct 2018. [Online]. Available: \url{http://arxiv.org/abs/1710.09435}
\BIBentrySTDinterwordspacing

\bibitem{Sportiello2011}
\BIBentryALTinterwordspacing
L.~Sportiello and S.~Zanero, ``File {Block} {Classification} by {Support} {Vector} {Machine},'' in \emph{2011 {Sixth} {International} {Conference} on {Availability}, {Reliability} and {Security}}.\hskip 1em plus 0.5em minus 0.4em\relax IEEE, Aug. 2011, pp. 307--312. [Online]. Available: \url{http://ieeexplore.ieee.org/document/6045955/}
\BIBentrySTDinterwordspacing

\bibitem{Xiong_bert_2024}
\BIBentryALTinterwordspacing
J.~Xiong, C.~Jiang, Z.~Zhao, Y.~Qiao, N.~Zhang, M.~Feng, and X.~Wang, ``Selecting the best fit software programming languages: Using bert for file format detection,'' \emph{Journal of Theory and Practice of Engineering Science}, vol.~4, no.~06, p. 20–28, Jul. 2024. [Online]. Available: \url{https://centuryscipub.com/index.php/jtpes/article/view/615}
\BIBentrySTDinterwordspacing

\bibitem{Schuster2012JapaneseAK}
\BIBentryALTinterwordspacing
M.~Schuster and K.~Nakajima, ``Japanese and korean voice search,'' \emph{2012 IEEE International Conference on Acoustics, Speech and Signal Processing (ICASSP)}, pp. 5149--5152, 2012. [Online]. Available: \url{https://api.semanticscholar.org/CorpusID:22320655}
\BIBentrySTDinterwordspacing

\bibitem{Nair2010}
V.~Nair and G.~E. Hinton, ``Rectified {Linear} {Units} {Improve} {Restricted} {Boltzmann} {Machines},'' \emph{Proceedings of the 27th International Conference on Machine Learning}, pp. 807--814, 2010.

\bibitem{Clevert2016}
\BIBentryALTinterwordspacing
D.-A. Clevert, T.~Unterthiner, and S.~Hochreiter, ``Fast and {Accurate} {Deep} {Network} {Learning} by {Exponential} {Linear} {Units} ({ELUs}),'' in \emph{Proceedings of the {International} {Conference} on {Learning} {Representations} ({ICLR})}, 2016, arXiv: 1511.07289. [Online]. Available: \url{http://arxiv.org/abs/1511.07289}
\BIBentrySTDinterwordspacing

\bibitem{bigcode}
\BIBentryALTinterwordspacing
``Bigcode,'' 2026. [Online]. Available: \url{https://huggingface.co/bigcode}
\BIBentrySTDinterwordspacing

\bibitem{Kocetkov2022TheStack}
D.~Kocetkov, R.~Li, L.~Ben~Allal, J.~Li, C.~Mou, C.~Muñoz~Ferrandis, Y.~Jernite, M.~Mitchell, S.~Hughes, T.~Wolf, D.~Bahdanau, L.~von Werra, and H.~de~Vries, ``The stack: 3 tb of permissively licensed source code,'' \emph{Preprint}, 2022.

\bibitem{lozhkov2024starcoder}
A.~Lozhkov, R.~Li, L.~B. Allal, F.~Cassano, J.~Lamy-Poirier, N.~Tazi, A.~Tang, D.~Pykhtar, J.~Liu, Y.~Wei, T.~Liu, M.~Tian, D.~Kocetkov, A.~Zucker, Y.~Belkada, Z.~Wang, Q.~Liu, D.~Abulkhanov, I.~Paul, Z.~Li, W.-D. Li, M.~Risdal, J.~Li, J.~Zhu, T.~Y. Zhuo, E.~Zheltonozhskii, N.~O.~O. Dade, W.~Yu, L.~Krauß, N.~Jain, Y.~Su, X.~He, M.~Dey, E.~Abati, Y.~Chai, N.~Muennighoff, X.~Tang, M.~Oblokulov, C.~Akiki, M.~Marone, C.~Mou, M.~Mishra, A.~Gu, B.~Hui, T.~Dao, A.~Zebaze, O.~Dehaene, N.~Patry, C.~Xu, J.~McAuley, H.~Hu, T.~Scholak, S.~Paquet, J.~Robinson, C.~J. Anderson, N.~Chapados, M.~Patwary, N.~Tajbakhsh, Y.~Jernite, C.~M. Ferrandis, L.~Zhang, S.~Hughes, T.~Wolf, A.~Guha, L.~von Werra, and H.~de~Vries, ``Starcoder 2 and the stack v2: The next generation,'' 2024.

\bibitem{Loshchilov2019}
\BIBentryALTinterwordspacing
I.~Loshchilov and F.~Hutter, ``Decoupled {Weight} {Decay} {Regularization},'' in \emph{International {Conference} on {Learning} {Representations} ({ICLR})}, 2019. [Online]. Available: \url{https://github.com/loshchil/AdamW-and-SGDW}
\BIBentrySTDinterwordspacing

\bibitem{Tompson2015}
\BIBentryALTinterwordspacing
J.~Tompson, R.~Goroshin, A.~Jain, Y.~LeCun, and C.~Bregler, ``Efficient object localization using {Convolutional} {Networks},'' in \emph{2015 {IEEE} {Conference} on {Computer} {Vision} and {Pattern} {Recognition} ({CVPR})}.\hskip 1em plus 0.5em minus 0.4em\relax IEEE, Jun. 2015, pp. 648--656, arXiv: 1411.4280v1. [Online]. Available: \url{http://ieeexplore.ieee.org/document/7298664/}
\BIBentrySTDinterwordspacing

\bibitem{Raff2020a}
\BIBentryALTinterwordspacing
E.~Raff and C.~Nicholas, ``A {Survey} of {Machine} {Learning} {Methods} and {Challenges} for {Windows} {Malware} {Classification},'' in \emph{{NeurIPS} 2020 {Workshop}: {ML} {Retrospectives}, {Surveys} \& {Meta}-{Analyses} ({ML}-{RSA})}, 2020, arXiv: 2006.09271. [Online]. Available: \url{http://arxiv.org/abs/2006.09271}
\BIBentrySTDinterwordspacing

\end{thebibliography}

\end{document}